# Autonomous Network Defence using Reinforcement Learning

Myles Foley
Imperial College London
m.foley20@imperial.ac.uk

Chris Hicks
The Alan Turing Institute
c.hicks@turing.ac.uk

Kate Highnam
The Alan Turing Institute, Imperial College London
k.highnam19@imperial.ac.uk

Vasilios Mavroudis
The Alan Turing Institute
vmavroudis@turing.ac.uk

## ABSTRACT

In the network security arms race, the defender is significantly disadvantaged as they need to successfully detect and counter every malicious attack. In contrast, the attacker needs to succeed only once. To level the playing field, we investigate the effectiveness of autonomous agents in a realistic network defence scenario. We first outline the problem, provide the background on reinforcement learning and detail our proposed agent design. Using a network environment simulation, with 13 hosts spanning 3 subnets, we train a novel reinforcement learning agent and show that it can reliably defend continual attacks by two advanced persistent threat (APT) red agents: one with complete knowledge of the network layout and another which must discover resources through exploration but is more general.

## 1 INTRODUCTION

Securing a computer network and its subsystems against attackers is a complex task that requires both the right combination of tools and expert knowledge [12]. At present, this task is usually still handled by human operators.

However, human involvement increases the operational costs and the response times, leaving these systems at risk. For instance, in 2020, attackers were estimated to spend a median of 24 days undetected inside a defensive environment [7]. During this time the attacker can further infiltrate, compromise, exfiltrate and perform other malicious activities within the network. Compared with an adversary, a defender typically faces increased complexity as they need to remove the threat whilst minimising operational disruption.

To advance autonomous defence, we investigate the application of reinforcement learning (RL). In recent years, RL has excelled in game-playing scenarios and has even exceeded human level ability (e.g., ATARI arcade games [11], Dota 2 [3]). Despite its reach potential, only a few works have sought to apply RL in the context of network security [4, 6, 9].

In this paper, we present a hierarchical RL agent for autonomous network defence. We demonstrate its success in winning the first CAGE environment challenge [1, 13]. The challenge uses the CybORG environment [14] to simulate a live computer network with several hosts and critical servers, as seen in Figure 1. This environment utilises a high-fidelity emulator, realised using virtual machines running on Amazon Web Services (AWS), to ensure that all available actions and states are realistic [14].

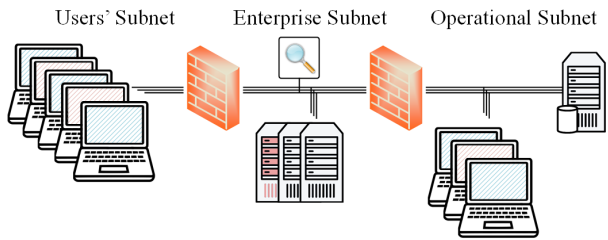

Figure 1: The CybORG network structure as specified in the environment of the CAGE Challenge [13].

Our contributions are as follows. We introduce our hierarchical agent architecture, featuring a controller agent selecting sub-agents which are trained against specific adversarial strategies. We demonstrate its capabilities against two adversaries: one which has prior knowledge of the network structure and another that does not. Through these experiments, we explain why the sum is greater than the parts. We publicly release our models' source code, the training setup, and trained models of our winning solution for researchers to compare against in the CAGE Challenge [1, 13].

This paper is structured in the following way. In Section 2 we describe the CybORG environment. We present our hierarchical solution and background in Section 3, followed by our results and a discussion in Sections 4 and 5. We conclude in Section 6.

## 2 THE CYBORG ENVIRONMENT

To showcase our model on the problem of network defence we leverage the recently proposed CybORG environment [14] as specified in the CAGE Challenge [13]. The same network structure is used as the agent's *environment* and is seen in Figure 1. This network delivers a contained, yet realistic setup, that can be emulated using Amazon Web Services (AWS) or simulated using Python. The network consists of nine users and four servers that are present across three subnets.

The game is turn-based between the attacker and defender. Each agent, which has limited visibility of the network state, agent chooses an *action* for each time-step. Based on the selected actions, the environment samples from a probability distribution (e.g. a valid node restoration may occasionally fail) to update its *state* and returns a *reward* to the agents. To prevent trivial solutions, the adversaries are given an initial foothold on a predetermined device in the users' subnet which cannot be 'restored' to a benign state by the defender.

The CAGE Challenge includes two adversaries: the B_lineAgent who has prior knowledge (i.e. perfect visibility of the network's structure but not of its current state), and the RedMeanderAgent who starts without prior information. Both agents share the same objective, to reach the operational server and, after escalating their privileges, disrupt its services. Due to its prior knowledge B_lineAgent follows an almost optimal exploitation trajectory to the operational server. In contrast, RedMeanderAgent scans the network for vulnerable hosts and gradually propagates through the subnets.

*State.* This is a vector of 52 bits, 4 bits for each host representing if the host has been scanned or exploited, and the access the attacker has on the host, none, user, administrator, or unknown.

*Action.* The defending agent performs actions at a host level: 1) Analysing processes on a specific host, 2) Terminating malicious processes, and 3) Restoring a host to a previous (benign) state. The defensive agent can also monitor network traffic or set up decoy host services. The attacking agents can: 1) Scan a subnet for hosts, 2) Scan the ports of a host, 3) Exploit a service on a port, 4) Escalate their access in a host, and 5) Disrupt the services on the operational server. Both agents have a 'sleep' action to perform no action on the network.

*Reward.* Defensive rewards are as follows: per admin access the red agent has, per turn: -0.1 for any host, -1 for any server. There is -10 for disruption on the operational server. Finally there is a -1 reward when any device must be restored. By having a negative reward for the defensive agent it forces the agent to minimise the effect of the attacker.

## 3 HIERARCHICAL RL LEARNING MODEL

RL is a form of machine learning that optimises a given reward, $r$. RL problems are formulated around an agent that exists in an environment, this agent takes an action, $a$, which changes the agents state, $s$, in the environment. The agent learns a policy $\pi$ that increases the long-term reward, accounted for by a discount factor $\gamma$, the *optimal* policy is one that maximises the reward, thus solving the task [15].

### 3.1 Proximal Policy Optimisation

Proximal Policy Optimisation (PPO) is an efficient policy gradient method for RL [11]. PPO can achieve human and super-human performance in a range of complex environments including 49 separate ATARI arcade games [11] and Dota 2 [3].

Taking a policy $\pi$ as $\pi_\theta$ (where $\theta \in \mathbb{R}$) define an objective function based on the expected total reward obtained from the environment: $J(\theta) = \mathbb{E}_{\pi_\theta}[\sum_{t=0}^{\infty} \gamma^t r_t]$. This can then be solved using an actor-critic architecture using a critic to measure how good an action is and an actor to select actions. This policy gradient is then: $\nabla_\theta J(\theta) = \mathbb{E}_{\pi_\theta}[\nabla_\theta \log \pi_\theta(s, a) A_{\pi_\theta}(s)]$.

Where $A_{\pi_\theta}(s) = Q_{\pi_\theta(s,a)} - V_{\pi_\theta}(s)$ is the advantage function to describe the advantage of action $a$ over the average action given by policy $\pi_\theta$ [15]. A deep neural-network can then approximate the state-value function $V_{\pi_\theta}(s)$ and another can model the policy $\pi_\theta$. PPO also introduces a clipping function in gradient descent.

This allows the model to balance large updates which lead to local optima and smaller updates which increase training time.

In our model we use the standard implementation of PPO included in RLLib [8] but modify the hyperparameters after a grid-search.

### 3.2 Curiosity

Curiosity promotes exploration in an environment via an intrinsic reward. Both sub agents (as defined in Section 3.3) use the Intrinsic Curiosity Module (ICM) proposed by [10], this incentivises agents to take actions where there is uncertainty in outcome, thus promoting the exploration to unknown states. ICM also reduces noise in this process by using only the relevant information in the state space. Again, we use the implementation in RLLib [8] modifying parameters after a grid-search. The effect of curiosity on these sub agents is illustrated clearly in the b-line-defence agent, where curiosity improves the reward by nearly double. ICM particularly overcomes the problem of developing strategies in the presence of randomness in the adversary and the possibility of actions being unsuccessful.

### 3.3 Hierarchy

To solve the problem as described in Section 2 we develop a hierarchical model that can defend against two different adversaries, the B_line and the Meander agents. In doing so we demonstrate a state-of-the-art solution to the network defence as in the CAGE Challenge.

For each of the sub agents in this hierarchy we train a PPO Agent with curiosity against a single adversary. We name these the b-line-defence (trained to defend against the B_line adversary), and the meander-defence (trained against the Meander adversary). Training occurs over *episodes* of 100 timesteps; the network and agents are then reset to their initial state. This continues until the reward of the defensive agent has converged.

The choice of PPO with curiosity was motivated by the improved performance in training as compared to APEX DQN [5], IMPALA [2], PPO [11], and PPO with curiosity [10]. This improved performance is seen as the ability to maximise the reward against the adversary, and thus minimising the effect of the adversary on the network.

We then train an RL agent 'controller', this deals with the high-level problem of identifying the adversary currently attacking the network. In training the controller defends each episode against a random adversary. The controller then selects one of the pre-trained sub-agents at each time-step to execute a low-level action. In this way the controller should identify the adversary and choose the agent trained to deal with the threat and mitigate it.

## 4 RESULTS

Here we present the results of our CAGE Challenge winning approach to autonomous network defence.

### 4.1 Challenge Evaluation

CAGE Challenge submissions were evaluated by testing the proposed defensive 'blue' agent against three unique attacking 'red' agents. The B_lineAgent, RedMeanderAgent, and SleepAgent are faced for 1000 episodes, for 30, 50, and 100 steps of the challenge environment. The defensive blue agent's score in each of the 9 different evaluation scenarios, represented by all combinations of red agents and episode step lengths, is averaged for the final score. The result of our winning submission in the CAGE Challenge is shown in Table 1.

| Red Agent | 30 Steps | 50 Steps | 100 Steps |
|---|---|---|---|
| B_lineAgent | -2.83 | -3.56 | -8.09 |
| RedMeanderAgent | -2.66 | -3.22 | -4.58 |
| SleepAgent | 0.0 | 0.0 | 0.0 |

Table 1: The results of our proposed solution to autonomous network defence in the CAGE challenge environment.

The SleepAgent only performs the Sleep action, therefore the perfect score, which our agent always achieves, is 0. Against the more challenging RedMeanderAgent and B_lineAgent our defensive agent maintains an average score of below -10 in all scenarios. This is particularly impressive when considering that a compromised operational server is penalised with a score of -10 for every step. On average, the operational server is never compromised when our defensive blue agent is protecting the CAGE Challenge environment.

### 4.2 Extended Evaluation

To help motivate, understand, and measure the performance of our hierarchical RL agent we perform an extended evaluation. Table 2 shows that whilst our B-line-defence and Meander-defence agents perform well against the B_lineAgent and RedMeanderAgent, respectively; the policies fail to generalise and the average performance of these agents in both tasks is unsatisfactory. Our Hierarchical agent performs very well at choosing the best specialised sub-agent for each observation. This is evidenced in comparison with a privileged Hierarchical-perfect agent and a Hierarchical-chance agent that chooses a blue agent uniformly at random. The Hierarchical -perfect agent has perfect knowledge of the current red agent adversary and always chooses the corresponding defensive blue agent. Our Hierarchical agent does far better than chance and scores higher than the sum of the B-line-defence and Meander-defence agents.

| Blue Agent | B_lineAgent | RedMeanderAgent | Avg. |
|---|---|---|---|
| B-line-defence | -10.80 | -61.05 | -35.93 |
| Meander-defence | -38.52 | -7.06 | -22.79 |
| Hierarchical | -11.01 | -5.84 | -8.43 |
| Hierarchical-perfect | -10.7 | -6.12 | -8.41 |
| Hierarchical-chance | -41.68 | -33.52 | -37.6 |

Table 2: The performance of each blue agent against the B_lineAgent and RedMeanderAgent adversaries.

## 5 DISCUSSION

The results show our hierarchical approach to autonomous network defence, which is state-of-the-art in the CAGE environment, outperforms the defensive capabilities of any sub-agent trained against a single red agent. We suggest that neither the B-line-defence or Meander-defence agent can fully generalise from training against a single adversary and that our hierarchical architecture provides a general mechanism to combine specialised sub-agents into a more widely applicable defensive capability. The theory that avoiding overfitting is critical to high-performance autonomous network defence is also supported by the relative performance of the B_lineAgent and RedMeanderAgent shown in Table 2. The B_lineAgent, which uses specialised knowledge of the network structure to execute a more efficient but less general attack, trains a worse-performing B-line-defence agent; in the average case, than the less specialised RedMeanderAgent. Additionally, our experiments with curiosity, a technique which encourages greater generalisation [10], substantially improves the performance of the B-line-defence agent even against the B_lineAgent itself.

## 6 CONCLUSION

This paper investigates the application of intelligent agents to autonomously defend a computer network. Utilising the recently proposed CAGE Challenge scenario, and CybORG (an autonomous network defence environment), we develop a hierarchical RL agent that can defend against multiple APT adversaries over varying lengths of time and overcomes the performance limitations of training against a single adversary. Surprisingly, we show that our hierarchy of specialised agents outperforms any of its individual components and provides for a more generalised defensive capability. We provide an open-source implementation of our state-of-the-art solution which we hope will help to fill the gap in autonomous network defence research. Promising directions for future research include adaptive fine-tuning of specialised agents by the controller, encouraging even greater generalisation, and formalising the principles which can ensure and quantify it, and multi-agent team defensive capabilities.

## ACKNOWLEDGEMENTS

The authors would like to acknowledge support from the Defence and Security Programme at The Alan Turing Institute, funded by the Government Communications Headquarters (GCHQ).